\documentclass[10pt,twocolumn,letterpaper]{article}

\usepackage{3dv}
\usepackage{times}
\usepackage{epsfig}
\usepackage{graphicx}
\usepackage{amsmath}
\usepackage{amssymb}
\usepackage{booktabs}

\usepackage{soul}

\usepackage[pagebackref=true,breaklinks=true,letterpaper=true,colorlinks,bookmarks=false]{hyperref}
\usepackage{appendix}

\def\aa{\mathbf{a}}
\def\bb{\mathbf{b}}
\def\cc{\mathbf{c}}
\def\xx{\mathbf{x}}
\def\yy{\mathbf{y}}

\def\ee{\mathbf{e}}
\def\ll{\mathbf{l}}
\def\pp{\mathbf{p}}

\def\mm{\mathbf{m}}
\def\uu{\mathbf{u}}
\def\cc{\mathbf{c}}
\def\ss{\mathbf{s}}

\threedvfinalcopy 

\def\threedvPaperID{126} 

\ifthreedvfinal\fi
\begin{document}

\title{Generalized Pose-and-Scale Estimation using 4-Point Congruence Constraints}

\author{Victor Fragoso\\
Microsoft\\
{\tt\small victor.fragoso@microsoft.com}
\and
Sudipta N. Sinha\\
Microsoft\\
{\tt\small sudipta.sinha@microsoft.com}
}

\maketitle

\begin{abstract}
We present gP4Pc, a new method for computing the absolute pose of a generalized camera with unknown internal scale from four corresponding 3D point-and-ray pairs. Unlike most pose-and-scale methods, gP4Pc is based on constraints arising from the congruence of shapes defined by two sets of four points related by an unknown similarity transformation. By choosing a novel parametrization for the problem, we derive a system of four quadratic equations in four scalar variables. The variables represent the distances of 3D points along the rays from the camera centers. After solving this system via Gr{\"o}bner basis-based automatic polynomial solvers, we compute the similarity transformation using an efficient 3D point-point alignment method.
We also propose a specialized variant of our solver for the case of coplanar points, which is computationally very efficient and about $3\times$ faster than the fastest existing solver. Our experiments on real and synthetic datasets, demonstrate that gP4Pc is among the fastest methods in terms of total running time when used within a RANSAC framework, while achieving competitive numerical stability, accuracy, and robustness to noise.
\end{abstract}

\section{Introduction}
\label{sec:introduction}

\let\thefootnote\relax\footnote{$^{1}$ This work was presented at IEEE 3DV 2020.}
\let\thefootnote\relax\footnote{$^{2}$ Code: \url{https://github.com/vfragoso/gp4pc}}

Absolute camera pose estimation from correspondences between 3D points and 2D image
coordinates (or viewing rays) is a fundamental task in computer vision. It has numerous applications in structure from motion (SfM)~\cite{sweeney2015theia,schoenberger2016sfm}, SLAM~\cite{orbslam}, image-based localization~\cite{lim2015real,sattler2011fast} for augmented reality, and robotics systems. While single camera pose estimation is a well studied topic~\cite{haralick1994,Lepetit2008},
recently, the topic of multi-camera pose estimation has been receiving considerable attention~\cite{camposeco2016,kneip2013,li2008,kneip2014upnp,ventura2014,nister2007,sweeney2014}.

\begin{figure}
\centering
\includegraphics[width=0.98\linewidth]{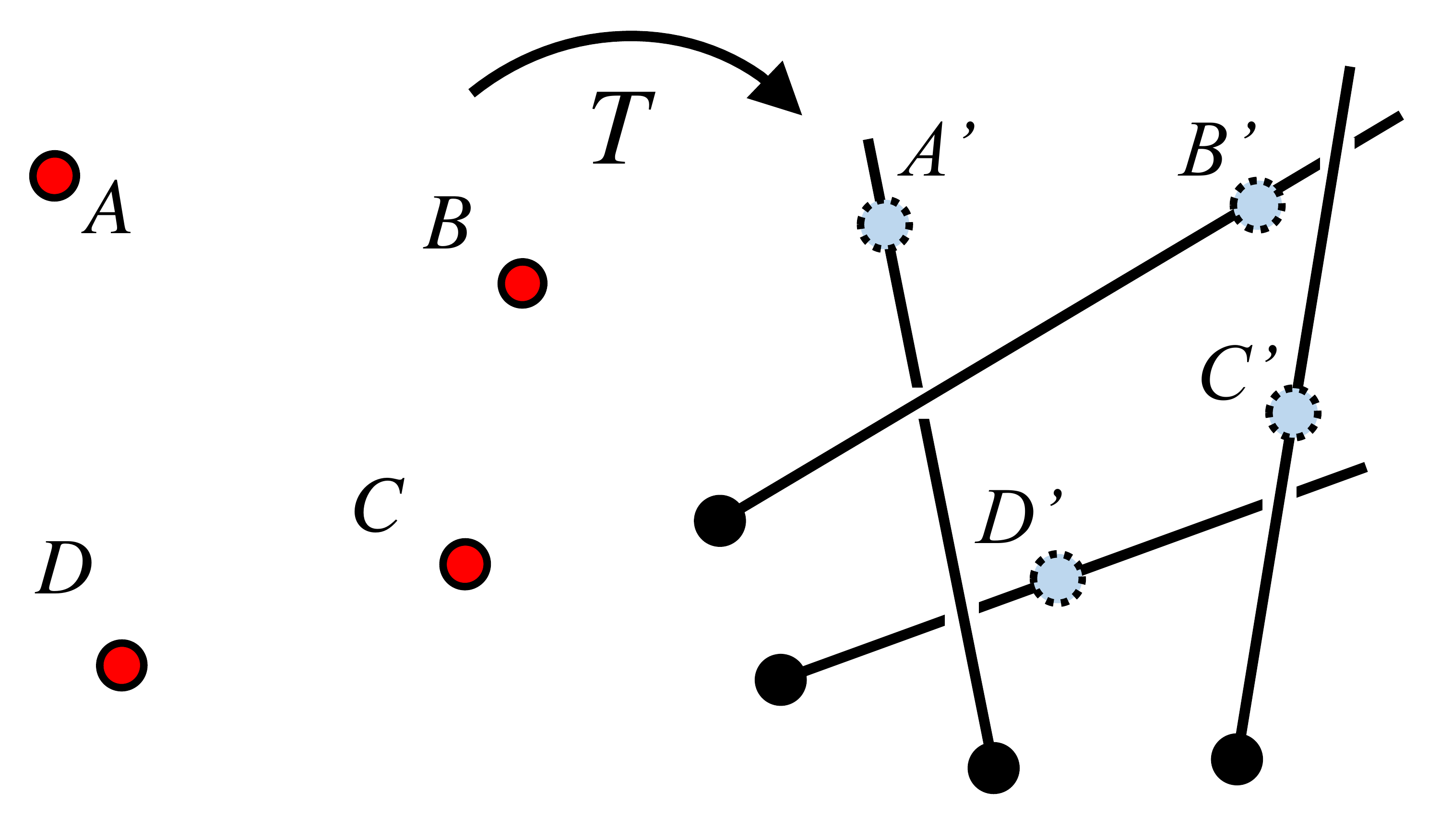}
\vspace{-2mm}
\caption{
gP4Pc is a new pose-and-scale minimal solver that uses point congruence constraints. Given four 3D point-ray correspondences, the points, denoted by $A$, $B$, $C$ and $D$ are related to the unknown points $A^\prime$, $B^\prime$, $C^\prime$ and $D^\prime$
on the rays by an unknown similarity transformation $T$. gP4Pc first solves the points $A^\prime$, $B^\prime$, $C^\prime$ and $D^\prime$ such that those points are congruent (or similar) to the original points and then computes $T$ from the point-point pairs.}
\label{figure1}
\end{figure}

The multi-camera pose estimation task involves localizing multiple images \wrt~an existing reconstruction in a single step and is closely related to the task of aligning two reconstructions with different scales. The underlying problem is referred to as the generalized pose-and-scale problem, where the goal is to find a similarity transformation between a set of 3D points and a set of 3D lines or rays, such that the transformed 3D points lie on the corresponding 3D lines. In the minimal case of this problem, four point-to-ray correspondences are required. In this paper, we present gP4Pc, a novel minimal solver for the generalized pose-and-scale problem. Unlike existing pose-and-scale solvers (\eg,~\cite{fragoso2020, sweeney2014, sweeney2016, ventura2014}) that minimize a least-squares cost function, gP4Pc exploits the congruence relations of the 3D points (see Figure~\ref{figure1}); the suffix `c' in gP4Pc refers to the idea of point congruence.

Constraints satisfied by congruent point sets have been used
for point-based scan registration~\cite{aiger2008,mohamad2014} but
their use in minimal problems in computer vision is not very common.
Bujnak et al.~\cite{bujnak2012,bujnak2008} and Zheng et al.~\cite{zheng2014general}
used them for estimating single camera pose with unknown focal length.
Camposeco et al.~\cite{camposeco2016} used such constraints for generalized camera pose estimation
but only for the specialized case where one point-point and two point-ray matches are given.

We show that these congruence properties also
lead to efficient generalized pose-and-scale solvers in the general setting.
Specifically, by choosing a novel parametrization, we derive a system of four quadratic equations
in four variables that we solve efficiently via Gr{\"o}bner-based polynomial solvers. These four variables
represent the distances from the pinholes or the camera projection centers to the 3D points along their associated viewing rays. Given the distances, we obtain the transformed 3D points in the generalized camera's coordinate frame and then compute the 3D similarity transformation from the four point pairs using
Umeyama's method~\cite{umeyama1991least}.
In contrast, most existing minimal solvers directly compute the similarity transformation.

\vspace{2mm}

\noindent \textbf{Contributions.}
\textbf{(1)}~We propose a new minimal solver for the generalized pose-and-scale estimation problem that is based on geometric relationships satisfied by two congruent 3D point sets up to unknown scale. Ours is the first general purpose solver that is derived from only congruence constraints.
\textbf{(2)}~We also exploit congruence constraints to derive a specialized solver for the case when the four 3D points are coplanar, which is very computationally efficient.


Our experiments on synthetic and real pose-and-scale estimation datasets show that gP4Pc has similar accuracy to the best performing methods such as gDLS~\cite{sweeney2014}, gDLS+++~\cite{sweeney2016}. However, gP4Pc leads to faster pose estimation than the aforementioned methods when using the minimal solvers within a RANSAC estimation framework. Finally, our specialized solver for coplanar points is about $3\times$ faster than existing methods and gives accurate results. 
\section{Related Work}
\label{sec:relatedwork}

Motivated by applications using multi-camera systems, Pless~\cite{pless2003using} first studied
generalized camera models where the viewing rays do not meet in a single
center of projection. Then, Nister~\cite{Nistr2004AMS} proposed the first
absolute pose estimation method for a generalized camera. Subsequently,
Ventura \etal~\cite{ventura2014} introduced the generalized pose-and-scale
problem for cameras where the internal scale is unknown
and proposed gP+s, the first pose-and-scale estimator. Since then, researchers proposed different pose-and-scale estimators ~\cite{camposeco2016,fragoso2020,kukelova2016,sweeney2014,sweeney2016} that improve speed and accuracy by integrating additional constraints (\eg, inertial sensors) or deriving efficient polynomial solvers. We review these methods for generalized cameras in the following sections.

\vspace{2mm}

\noindent \textbf{Absolute pose minimal solvers.}
There is extensive work on camera pose estimation, especially for pinhole cameras~\cite{haralick1994,ransac,ke2017efficient,Lepetit2008,hesch2011,zheng2013revisiting,persson2018lambda}. These efforts focus mostly on minimal solvers for the three point case, which are efficient and easy to use within
robust estimation frameworks such as RANSAC~\cite{ransac}.
For generalized cameras, Nister~\cite{Nistr2004AMS} and then Nister and Stewenius~\cite{nister2007}
presented the first pose estimation methods. They studied generalized cameras with known scale, \ie, assumed the distance between the multiple projection centers to be known. Other related works are those of Chen and Chang~\cite{chen2004pose}, Lee \etal~\cite{lee2016minimal}, Schweighofer and Pinz~\cite{schweighofer2006}, Kneip \etal~\cite{kneip2013} and
Fragoso \etal~\cite{fragoso2020}.

\vspace{2mm}

\noindent \textbf{gP+s.} Ventura~\etal~\cite{ventura2014} introduced the generalized pose-and-scale problem for generalized cameras with unknown scale, \ie, when the distance between the multiple projection centers is unknown. Their method solves the rotation, translation, and scale directly by solving a polynomial system encoding the null space of a linear system which describes the solution space of the unknown parameters. They solve this polynomial system using automatic Gr{\"o}bner basis-based polynomial solvers~\cite{kukelova2008} and handle both minimal and overdetermined problems.

\vspace{2mm}

\noindent \textbf{gDLS.}
Sweeney \etal~\cite{sweeney2014} presented gDLS, a pose-and-scale estimator for a generalized camera~\cite{pless2003using}
inspired by the work of Hesch~\etal~\cite{hesch2011}.
gDLS frames the pose-and-scale problem as a least squares problem. They show that it is possible to derive linear
relationships between scale, depths (distance from camera center to a 3D point) and translation as a function of
rotation. gDLS exploits these linear relationships to rewrite the gDLS least-squares problem as a function
of only the rotation. The least squares solution can be found by solving a polynomial system in the Cayley-Gibbs-Rodrigues
rotation parameters, using the Macaulay-matrix-based polynomial solver. The solutions
correspond to all the critical points of the least-squares objective.

\vspace{2mm}

\noindent \textbf{uPnP.} Kneip~\etal~\cite{kneip2014upnp} presented uPnP, which works for both single and generalized cameras. Similar to DLS~\cite{hesch2011}, uPnP frames pose estimation as a least-squares problem and rewrites it as a function of only rotational parameters. While DLS and gDLS use a Cayley-Gibbs-Rodrigues rotation parametrization, uPnP uses quaternions. To solve for the quaternions, Kneip~\etal~ find all the critical points of the least-squares function by solving the polynomial system in the quaternion parameters using a fast and universal Gr{\"o}bner-based polynomial solver. Unlike gDLS, uPnP cannot recover the scale of a generalized camera.

\vspace{2mm}

\noindent \textbf{gP1R2+s.} Camposeco \etal~\cite{camposeco2016} proposed a specialized solver for the
generalized pose-and-scale problem that assumes a specific form of input -- two point--rays and one 3D point--point
correspondence. Their method is very efficient and accurate and can be used in SLAM applications where mixed
point and ray correspondences are available.

\vspace{2mm}

\noindent \textbf{gDLS+++.} Sweeney \etal~\cite{sweeney2016} proposed a faster version of gDLS that uses
the same polynomial solver as uPnP~\cite{kneip2014upnp} and they extended that method to
handle unknown scale.

\vspace{2mm}

\noindent \textbf{3Q3} Kukelova \etal ~\cite{kukelova2016}.
proposed a general method to solve a polynomial system of three quadrics. Their method can be
used to solve the pose problem for which they can solve for the roots of the octic polynomial very efficiently
and avoid computing a Gr{\"o}bner basis.

\vspace{2mm}

\noindent \textbf{Relation to gDLS}~\cite{sweeney2014}. Different from methods such as gDLS that use a least-squares formulation, our proposed method (gP4Pc)
uses pure geometric constraints and the polynomial system at the core of our method does not involve the scale, translation or rotation parameters.
Instead, our formulation exploits point congruence relations to first estimate the position of the four points along the viewing rays.
Our polynomial system is numerically stable and achieves a comparable speed to gDLS. After computing the distances, we estimate the similarity transformation using a 3D point-point registration method~\cite{horn1987closed,umeyama1991least}. Unlike gDLS that can struggle with cases where no rotations exist because of its Cayley-Gibbs-Rodrigues rotation parameterization, gP4Pc works in these cases because it solves for rotation as part of the point-to-point alignment step.

\vspace{2mm}


\noindent \textbf{Relation to gP1R2+s~\cite{camposeco2016} and P4Pf methods}.
Camposeco \etal~\cite{camposeco2016} solves the pose-and-scale problem using constraints derived from triangle congruence
and a parametrization similar to ours. However, they proposed a specialized
solver that uses one point-point and two point-ray correspondence, whereas we handle the general case of four
point-ray correspondences. Specifically, they derive distance constraints from a triangle
formed by one known point and two unknown points whereas we derive the constraints from four unknown points.
Bujnak et al.~\cite{bujnak2012,bujnak2008} and Zheng et al.~\cite{zheng2014general} used distance ratio
constraints for the P4Pf and PnPf problems respectively. This involves estimating single camera pose with unknown focal length
in the minimal and non-minimal settings. While some of our derived constraints are similar, they are used in a different problem.

\vspace{2mm}

\noindent \textbf{Relation to 4PCS registration methods.}
The notion of congruence and affine invariance in point sets are well studied and used in previous work on point-based scan registration~\cite{huttenlocher1991,aiger2008,mohamad2014,theiler2014,theiler2014fast}.
We were inspired by the 4PCS method and its variants, but we use congruence properties
in a completely different way from prior works by using them to derive algebraic constraints in our problem. 
\section{Key Elements of Proposed Method}
\label{sec:method}
In this section, we first review affine invariance and congruence properties of 4-point sets. Then, we describe the
new geometric constraints and the derivation of our method.

\subsection{4-Point Congruent Sets}
\label{sec:congruent_sets}

\begin{figure}
\centering
\includegraphics[width=\linewidth]{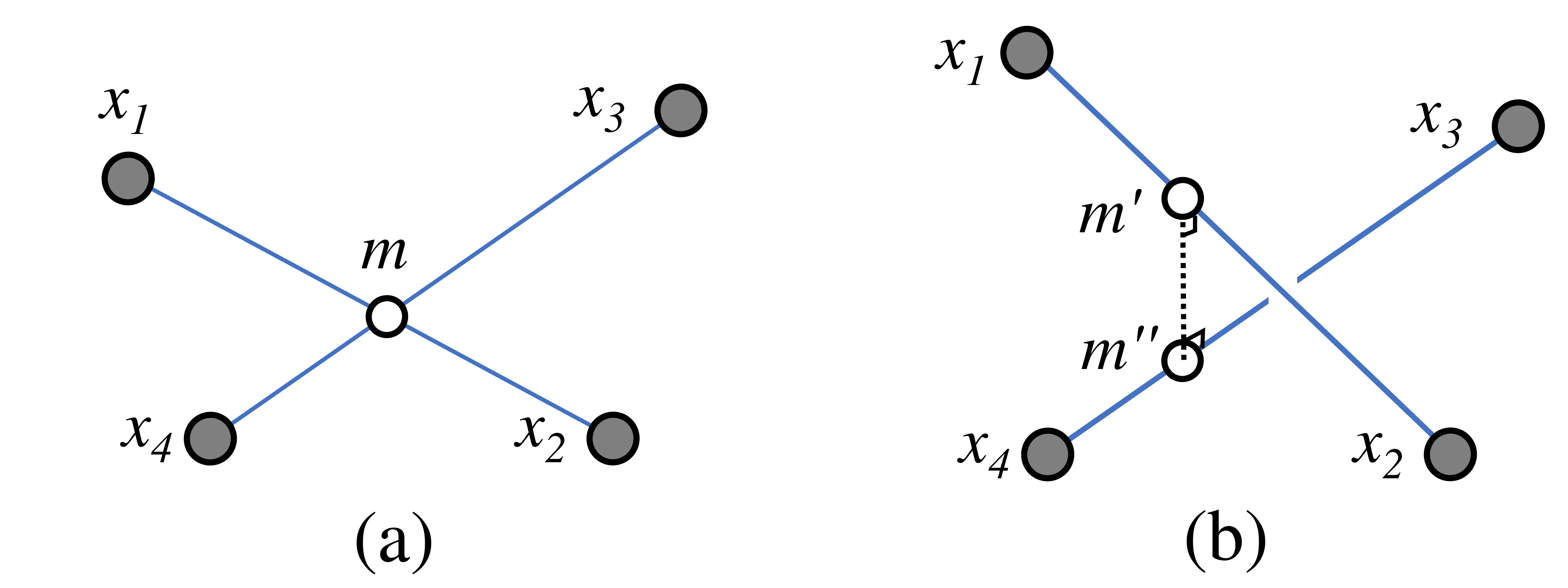}\\
\caption{(a) Four coplanar points $\xx_1$, $\xx_2$, $\xx_3$ and $\xx_4$ and the two lines intersecting at the point $\mathbf{m}$.
(b) For non-coplanar points, the closest points on the two lines are $\mm^\prime$ and $\mm^{\prime\prime}$ respectively. The line joining $\mm^\prime$ and $\mm^{\prime\prime}$ is orthogonal to the two lines.}
\label{figure2}
\end{figure}

Given three collinear points $\aa, \bb, \cc$, the ratio $\frac{||\aa-\bb||}{||\aa-\cc||}$
is preserved under all affine transformations. Huttenlocher~\cite{huttenlocher1991} used
this invariant to find all sets of four 2D points in the plane that are equivalent under
affine transforms. This property also holds for affine transformations in $\mathbb{R}^3$~\cite{aiger2008} which is useful in our case.

Let $\mm$ denote the point where the two line segments $\overline{\xx_1\xx_2}$ and $\overline{\xx_3\xx_4}$ intersect, where $\xx_1,\xx_2,\xx_3$, and $\xx_4$ are four coplanar points which are not all collinear; there is always a way to choose the pairs such that the lines intersect. Then, the two ratios $r_1$ and $r_2$ can be defined as follows:
\begin{equation}
r_1 = \frac{\| \xx_1 - \mm\|}{\|\xx_1 - \xx_2\|},
\,\,\,\,\,\,  r_2 = \frac{\|\xx_3 - \mm\|}{\|\xx_3 - \xx_4\|}.
\label{ratio-coplanar}
\end{equation}
These ratios for non-coplanar points are defined as follows:
\begin{equation}
r_1 = \frac{||\xx_1 - \mm^\prime||}{||\xx_1 - \xx_2||},
\,\,\,\,\,\,  r_2 = \frac{||\xx_3 - \mm^{\prime\prime}||}{||\xx_3 - \xx_4||},
\label{ratio-general}
\end{equation}
where the two points $\mm^\prime$ and $\mm^{\prime\prime}$ lie on the lines $\overline{\xx_1\xx_2}$
and $\overline{\xx_3\xx_4}$, respectively, such that the line connecting $\mm^\prime$ and $\mm^{\prime\prime}$
is orthogonal to both $\overline{\xx_1\xx_2}$ and $\overline{\xx_3\xx_4}$. See Fig.~\ref{figure2} for an illustration of both geometric settings.

Aiger~\etal~\cite{aiger2008} proposed the 4PCS method to register two 3D point scans, where they compute ratios from four coplanar point base set defined in Equation~\eqref{ratio-coplanar} in the first scan, and then efficiently find all subsets of four points in the second scan that are approximately congruent to the base set. Mohamad~\etal~\cite{mohamad2014} generalized the 4PCS algorithm for non-coplanar four point bases by computing the ratios described in Equation~\eqref{ratio-general}.
Previously, the congruence properties described here were used to
efficiently search for corresponding points in various point set registration tasks~\cite{aiger2008,huttenlocher1991,mohamad2014}. In contrast, we use them to derive minimal solvers for 3D point-to-ray registration problems.

\subsection{Constraints from affine invariants}

In the minimal setting, we are given four 3D points $\xx_1, \xx_2, \xx_3, \xx_4$ in one coordinate frame
and four corresponding 3D rays $\ll_1, \ll_2, \ll_3, \ll_4$  in a second
coordinate frame. By parameterizing the 3D points on each ray $\ll_i$ using a 3D point $\pp_i$
(the pinhole) and a 3D unit vector $\uu_i$ directed towards the scene, the points on the ray in front of the camera can be expressed as $\pp_i + s_i \uu_i$, where $s_i \ge 0$, and $\xx_i, \pp_i, \uu_i \in \mathbb{R}^3$.

\begin{figure}
\centering
\includegraphics[width=0.92\linewidth]{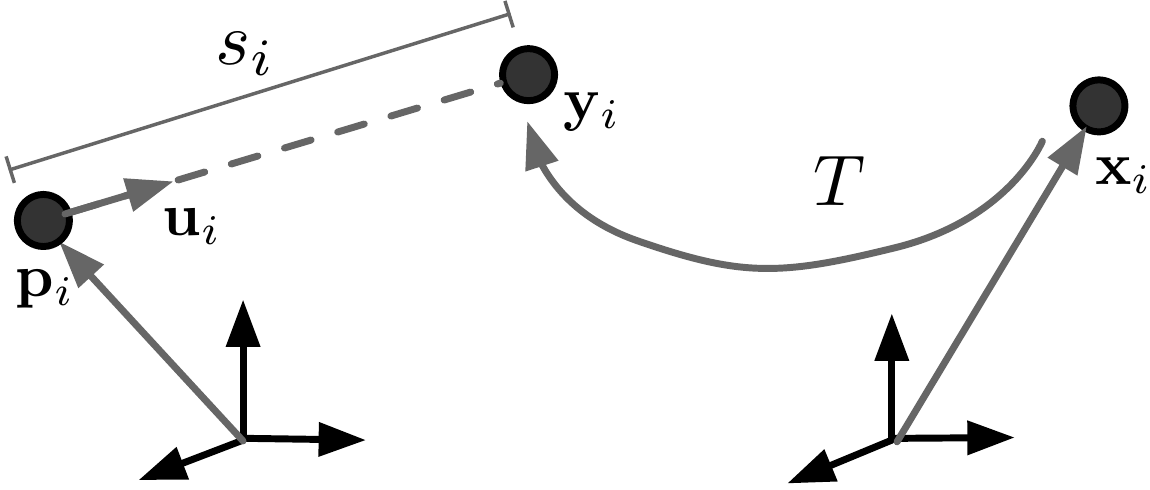}
\caption{{\bf Point-ray correspondence.} The similarity transformation $T$ maps the point $\mathbf{x}_i$ into a point $\mathbf{y}_i = \mathbf{p}_i + s_i \mathbf{u}_i$, where $\mathbf{p}_i$ is the pinhole, $s_i > 0$ is a scalar, and $\mathbf{u}_i$ is the line-unit vector.}
\label{fig:point_to_line}
\end{figure}

The unknown similarity transformation $T$ maps each point $\xx_i$ from
the first coordinate frame into a point $\yy_i$ in the second coordinate frame,
such that the point $\yy_i$ lies on the corresponding ray $\ll_i$ (see Figure~\ref{fig:point_to_line}).
Therefore, there exists four scalars $s_1, s_2, s_3, s_4$ such that
\begin{eqnarray}
\yy_i = \pp_i + s_i \uu_i, \quad i\,\in\,\{1,\,2,\,3,\,4\}.
\label{parameterization}
\end{eqnarray}

We now use the congruence relations (see Sec.~\ref{sec:congruent_sets}) satisfied by the four points $\yy_i$ for $i= 1,2,3,4$ to derive new constraints for solving the pose-and-scale problem. These constraints provide us
with linear and nonlinear equations in the four unknowns, \ie, $s_1, s_2, s_3$ and $s_4$, respectively.
In our method, we first directly solve for $s_1, s_2, s_3$ and $s_4$ without needing to deal with the
scale, rotation and translation parameters of the similarity transformation $T$. Then, substituting the
solutions into Equation~\eqref{parameterization}, we obtain the coordinates of $\yy_i$ for $i= 1,2,3,4$.
Finally, we estimate the similarity transformation from the four corresponding point pairs
$\{\xx_i \leftrightarrow \yy_i\}$ for $i= 1,2,3,4$, respectively.

\paragraph{Case 1: Coplanar Points.}
Given the ratios $r_1$ and $r_2$ obtained from the input points $\xx_1, \xx_2, \xx_3, \xx_4$ according to
Equation~\eqref{ratio-coplanar}, we can define two intermediate points for the corresponding
lines $\overline{\yy_1\yy_2}$ and $\overline{\yy_3\yy_4}$, yielding the following:
\begin{eqnarray}
\label{mm1234a}  \mm_{12} &=& (1 - r_1) \yy_1 + r_1 \yy_2\\
  \mm_{34} &=& (1 - r_2) \yy_3 + r_2 \yy_4.
\label{mm1234}
\end{eqnarray}

Based on the congruence property exploited by prior methods~\cite{aiger2008,huttenlocher1991},
we known that $\mm_{12} = \mm_{34}$.
After substitutions and rearranging variables, we get three linear equations in
$s_1$, $s_2$, $s_3$ and $s_4$ of the form:
\begin{eqnarray}
\nonumber
(1 - r_1) (\pp_1 + s_1 \uu_1) + r_1 (\pp_2 + s_2 \uu_2) = \\
(1 - r_2) (\pp_3 + s_3 \uu_3) + r_2 (\pp_4 + s_4 \uu_4).
\label{eq:linearequations}
\end{eqnarray}

\paragraph{Case 2: Non-coplanar Points.}
When the four points are non-coplanar, the associated intermediate points will not coincide.
However, as mentioned before, the line joining them will be orthogonal to the two
underlying  lines, $\overline{\yy_1\yy_2}$ and $\overline{\yy_3\yy_4}$, respectively. Thus, we have
\begin{eqnarray}
\label{eq:quadraticequations1and2a}
 (\yy_1 - \yy_2)^\intercal(\mm_{12}-\mm_{34}) &= 0\\
 (\yy_3 - \yy_4)^\intercal(\mm_{12}-\mm_{34}) &= 0.
\label{eq:quadraticequations1and2}
\end{eqnarray}
By substituting $\mm_{12}$ and $\mm_{34}$ from Equations~\eqref{mm1234} and $\yy_1, \yy_2, \yy_3$
and $\yy_4$ from Equation~\eqref{parameterization}, we get two quadratic equations in the
variables, $s_1$, $s_2$, $s_3$, and $s_4$.

\subsection{Constraints from ratios of distances}

\begin{figure}
\centering
\includegraphics[width=0.7\linewidth]{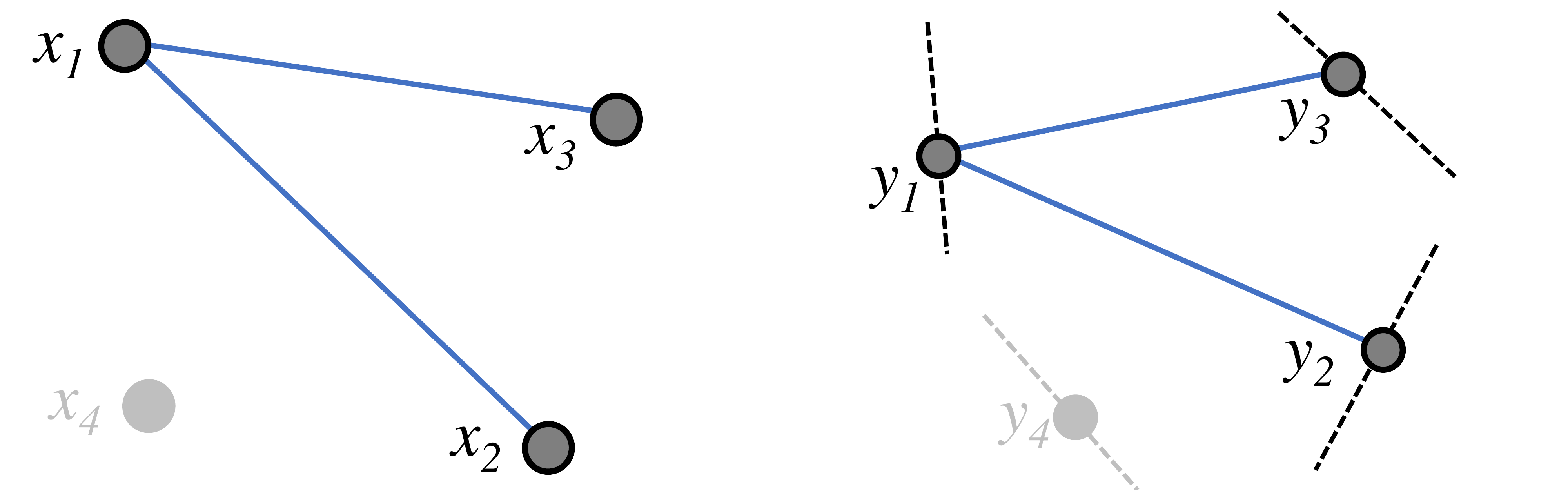}\\
(a)\hspace{3cm}(b)
\caption{\textbf{Congruent point sets and distance ratios.} (a) The four points $\xx_1$, $\xx_2$, $\xx_3$, $\xx_4$, with two line segments highlighted. (b) The congruent points $\yy_1$, $\yy_2$, $\yy_3$, $\yy_4$ and the corresponding line segments. Congruence implies that the length ratio of segments $\xx_1\xx_2$ and $\xx_1\xx_3$ is equal to that of $\yy_1\yy_2$ and $\yy_1\yy_3$.}
\label{fig:distanceratio}
\end{figure}

Consider the tetrahedra $A$ and $B$, formed by the point sets, $\{\xx_1, \xx_2, \xx_3, \xx_4\}$
and $\{\yy_1, \yy_2, \yy_3, \yy_4\}$, respectively. By definition, $A$ and $B$ must be congruent
to each other. Let us now consider a pair of edges in tetrahedron $A$, \eg, $\{\overline{\xx_1\xx_2},\overline{\xx_1\xx_3}\}$.
The corresponding edge pair in $B$ is $\{\overline{\yy_1\yy_2},\overline{\yy_1\yy_3}\}$ (see Figure~\ref{fig:distanceratio}).
Observe that due to the underlying congruence, the ratio of the length of the two edges in these
pairs must be the equal. Using $d(a, b)$ to denote the distance between points
$a$ and $b$, we have,
\begin{equation}
\frac{d(\xx_1, \xx_2)}{d(\xx_3,\xx_4)} = \frac{d(\yy_1, \yy_2)}{d(\yy_3,\yy_4)}.
\label{eq:distanceratio}
\end{equation}

Squaring both sides of the above equation and replacing the left side by $K_{1234}$  (constant term since $\xx_1, \xx_2, \xx_3$ and $\xx_4$ are known) gives the following:
\begin{equation}
d(\yy_1, \yy_2)^2 - K_{1234} d(\yy_3,\yy_4)^2 = 0.
\end{equation}
Rewriting $d(\yy_i, \yy_j)^2$ as $\ee_{ij}^\intercal\ee_{ij}$, where $\ee_{ij}$ is the vector between $\yy_i$
and $\yy_j$, we get
\begin{equation}
\ee_{12}^\intercal \ee_{12} - K_{1234} \ee_{34}^\intercal \ee_{34} = 0.
\label{nonlin_eq}
\end{equation}

After substituting $\yy_1, \yy_2, \yy_3$, and $\yy_4$ from Equation~\eqref{parameterization}, $\ee_{12}$
and $\ee_{34}$ has the following form.
\begin{align}
\ee_{12}&= \yy_1 - \yy_2 = \pp_5 + s_1 \uu_1 - s_2 \uu_2\\
\ee_{34}&= \yy_3 - \yy_4 = \pp_6 + s_3 \uu_3 - s_4 \uu_4.
\end{align}
In the equations above, we substituted the terms $\pp_1 - \pp_2$ and
$\pp_3 - \pp_4$ with two new terms
$\pp_5$ and $\pp_6$ respectively to simplify the notation. We can now rewrite $\ee_{12}^\intercal \ee_{12}$
as a quadratic polynomial $q$ in $s_1$, $s_2$, $s_3$ and $s_4$ as follows:
\begin{equation}
  q_a = s_1^2 + s_2^2 - 2 \uu_1^\intercal \uu_2 s_1s_2 + 2 \uu_1^\intercal \pp_5 s_1 - 2 \uu_2^\intercal \pp_5 s_2 + \pp_5^\intercal \pp_5.
  \label{poly12}
\end{equation}
Similarly, for $\ee_{34}^\intercal \ee_{34}$, we have the following polynomial:
\begin{equation}
  q_b = s_3^2 + s_4^2 - 2 \uu_3^\intercal \uu_4 s_3s_4 + 2 \uu_3^\intercal \pp_6 s_3 - 2 \uu_4^\intercal \pp_6 s_4 + \pp_6^\intercal \pp_6.
  \label{poly34}
  \vspace{2mm}\\
\end{equation}
Next, we arrange the 15 monomials in a column vector\\
$\ss = [s_1^2 \: s_2^2 \: s_3^2 \: s_4^2 \: s_1s_2 \: s_1s_3 \: s_1s_4 \: s_2s_3 \: s_2s_4 \: s_3s_4 \: s_1 \: s_2 \: s_3 \: s_4 \: 1]^\intercal$ and substitute the polynomials from Eqns.~\eqref{poly12} and~\eqref{poly34} into
Eqn.~\eqref{nonlin_eq} to get the following equation:
\begin{equation}
(\boldsymbol{\beta}_{12} - K_{1234} \boldsymbol{\beta}_{34})^\intercal \ss = 0,
\label{eq:1234}
\end{equation}
where $\boldsymbol{\beta}_{12}$ and $\boldsymbol{\beta}_{34}$ are column vectors denoting polynomials coefficients from
Eqns.~\ref{poly12} and~\ref{poly34}, respectively.

Given six edges, there are fifteen distance ratios for all the unique edge pairs.
However, only five of these are independent and the remaining ten can be derived from the five. Without
loss of generality we select five pairs: $(\ee_{12}, \ee_{34})$, $(\ee_{12}, \ee_{13})$, $(\ee_{12}, \ee_{14})$,
$(\ee_{12}, \ee_{23})$, and $(\ee_{12}, \ee_{24})$. Other than Equation~\eqref{eq:1234}, we now have four
other equations encoding the distance ratio constraints:
\begin{equation}
\left\{ \begin{array}{l}
(\boldsymbol{\beta}_{12} - K_{1213} \boldsymbol{\beta}_{13})^\intercal \ss = 0  \\
(\boldsymbol{\beta}_{12} - K_{1214} \boldsymbol{\beta}_{14})^\intercal \ss = 0  \\
(\boldsymbol{\beta}_{12} - K_{1223} \boldsymbol{\beta}_{23})^\intercal \ss = 0  \\
(\boldsymbol{\beta}_{12} - K_{1224} \boldsymbol{\beta}_{24})^\intercal \ss = 0
\end{array}
\right.,
\label{eq:distanceratioequations}
\end{equation}
where $K_{ijkl}$ is the ratio of the squared lengths of edges $\ee_{ij}$ and $\ee_{kl}$ and the $\boldsymbol{\beta}_{ij}$'s are defined similarly
to $\boldsymbol{\beta}_{12}$ and $\boldsymbol{\beta}_{34}$.
Thus, the polynomial system encoding the congruency constraints is comprised of Equation~\eqref{eq:1234} and Equations~\eqref{eq:distanceratioequations}.

\section{Proposed Solver (gP4Pc)}
Our solver consists of two phases. The first one solves for $s_1$, $s_2$, $s_3$, and $s_4$. These four values encode the ``depths'' or distances between each point and the camera center corresponding to a point-to-ray pair. Then, the second phase computes the similarity transformation by aligning the four input points $\xx_1$, $\xx_2$, $\xx_3$, $\xx_4$ and their corresponding estimated points $\yy_1$, $\yy_2$, $\yy_3$, $\yy_4$ using Equation~\eqref{parameterization} and $s_1$, $s_2$, $s_3$, and $s_4$. The steps of our solution are the following:
\begin{enumerate}\itemsep1pt
\item Given the four input points $\xx_1$, $\xx_2$, $\xx_3$, $\xx_4$, find the closest points on lines $\overline{\xx_1\xx_2}$ and $\overline{\xx_3\xx_4}$ and then compute ratios $r_1$ and $r_2$ according to Equation~\eqref{ratio-general} and the terms $K_{ijkl}$ according to Equations~\eqref{eq:1234} and \eqref{eq:distanceratioequations}.
\item Compute the polynomial coefficients $\boldsymbol{\beta}_{ij}$ according to Equations~\eqref{eq:quadraticequations1and2a},~\eqref{eq:quadraticequations1and2},~\eqref{eq:1234} and~\eqref{eq:distanceratioequations} (we only use the first of the four equations in~\eqref{eq:distanceratioequations}).
\item Solve the polynomial system using code generated by an automatic solver generator~\cite{larsson2017,larsson2018beyond}.
\item Keep solutions that satisfy $s_i \geq 0 ~~\forall i=1,2,3,4$.
\item For each solution, compute $\yy_1$, $\yy_2$, $\yy_3$ and $\yy_4$ using Equation~\eqref{parameterization} and then compute the coordinate transformation from the point pairs -- ($\xx_1$,$\yy_1$), ($\xx_2$,$\yy_2$), ($\xx_3$,$\yy_3$) and ($\xx_4$,$\yy_4$) (see details in the next section).
\end{enumerate}

The order of the points and rays in the input to the polynomial solver matters at multiple steps. Initially, it matters when computing the ratios $r_1$, $r_2$, $K_{ijkl}$, and  when computing the coefficients of the Equations~\eqref{eq:distanceratio} and \eqref{eq:distanceratioequations}.
Nevertheless, thanks to the randomization inherent in RANSAC, several permutations are sampled. This mitigates the need to find a procedure to compute an optimal point permutation. We experimented with a variant of gP4Pc that uses six different permutations and combines all the solutions to obtain a larger pool of hypotheses. However, this variant did not outperform the simpler and efficient version that uses one permutation per minimal problem; see Sec.~\ref{sec:experiments}.

\subsection{Specialized method for coplanar points.}
Recall that when the input points are coplanar, there are three linear constraints in $s_1$, $s_2$, $s_3$ and $s_4$ (see Equations~\eqref{eq:linearequations}). However, we need another constraint to find a unique solution to the four unknowns.
So we use one of the quadratic equations encoding the distance ratio constraints (in our implementation, we always used Equation~\eqref{poly34}) associated with edge pair $(\ee_{12}, \ee_{13})$. Using the linear constraints, we can eliminate $s_1$, $s_2$, $s_3$ from the quadratic equation to obtain a new
quadratic equation in only $s_4$ which can have two real roots.
We keep the roots that are positive and backsubstitute the value of $s_4$ to obtain up to two solutions,
retaining only the solutions where all four values are real and positive.
The details are in the appendix. 
\section{Experimental Results}
\label{sec:experiments}

\begin{figure*}[t]
\centering
\includegraphics[width=\textwidth]{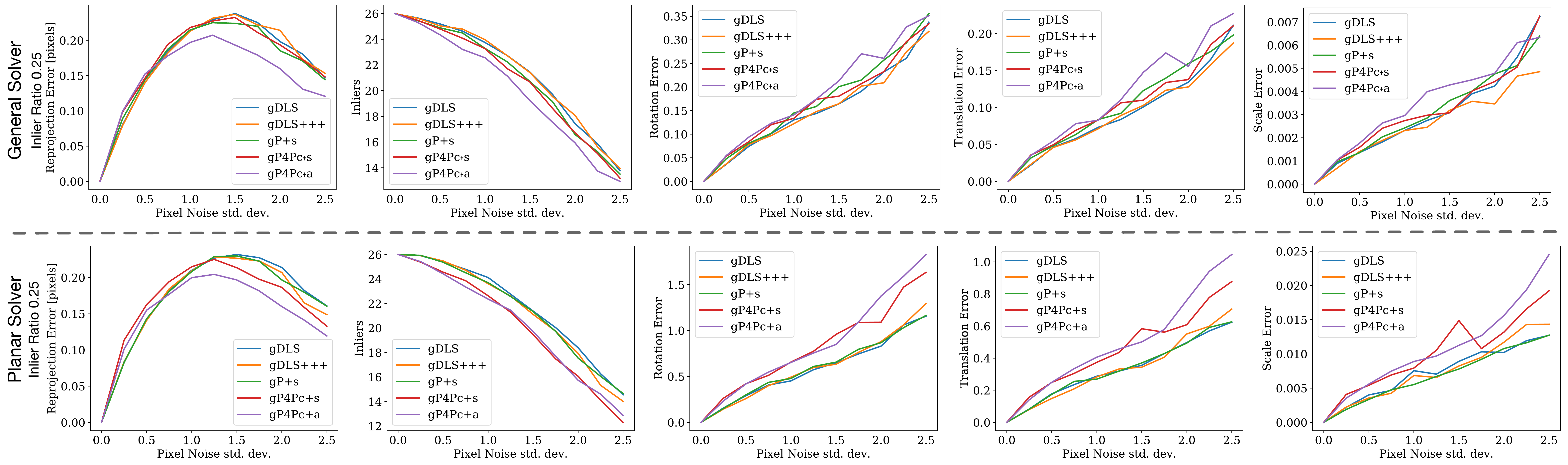}
\caption{{\bf Evaluation on synthetic data (with RANSAC loop)}. [Left to right] The effect of noise on average reprojection error, inlier count, rotation error, translation error and scale error respectively for different minimal solvers used within a RANSAC framework.}
\label{fig:synthetic_ransac}
\end{figure*}

\begin{figure}
\centering
\includegraphics[width=0.325\columnwidth]{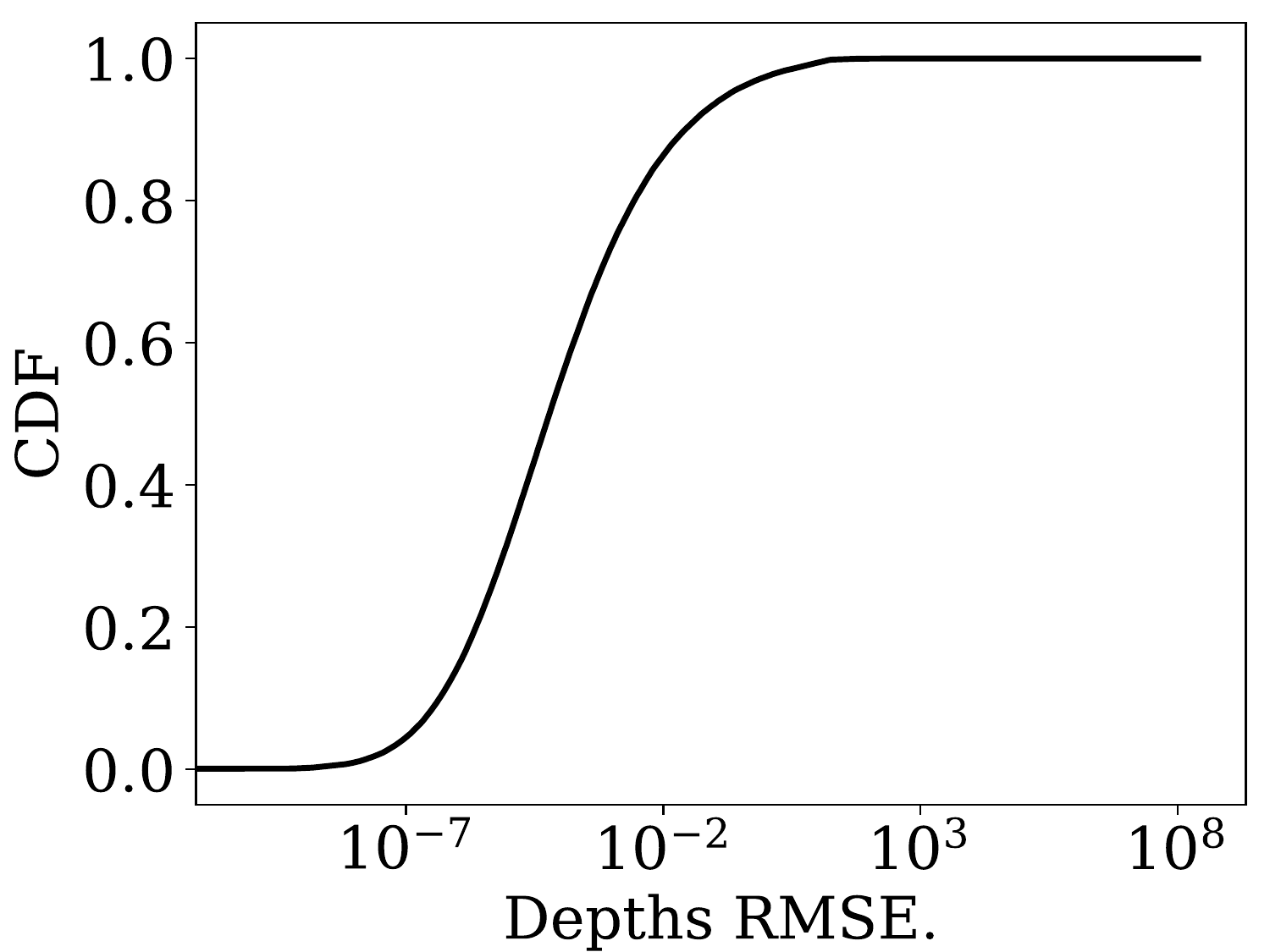}
\includegraphics[width=0.325\columnwidth]{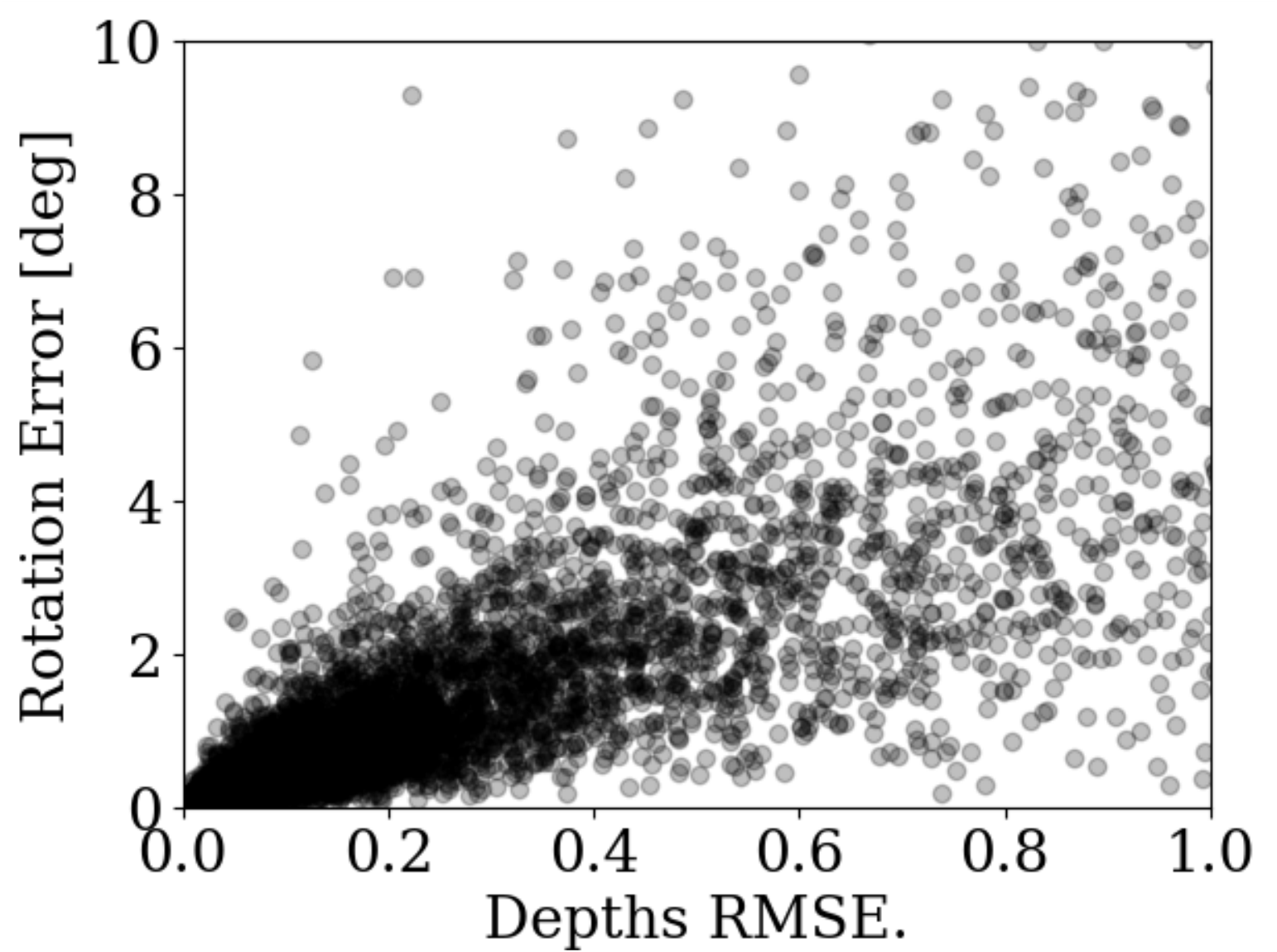}
\includegraphics[width=0.325\columnwidth]{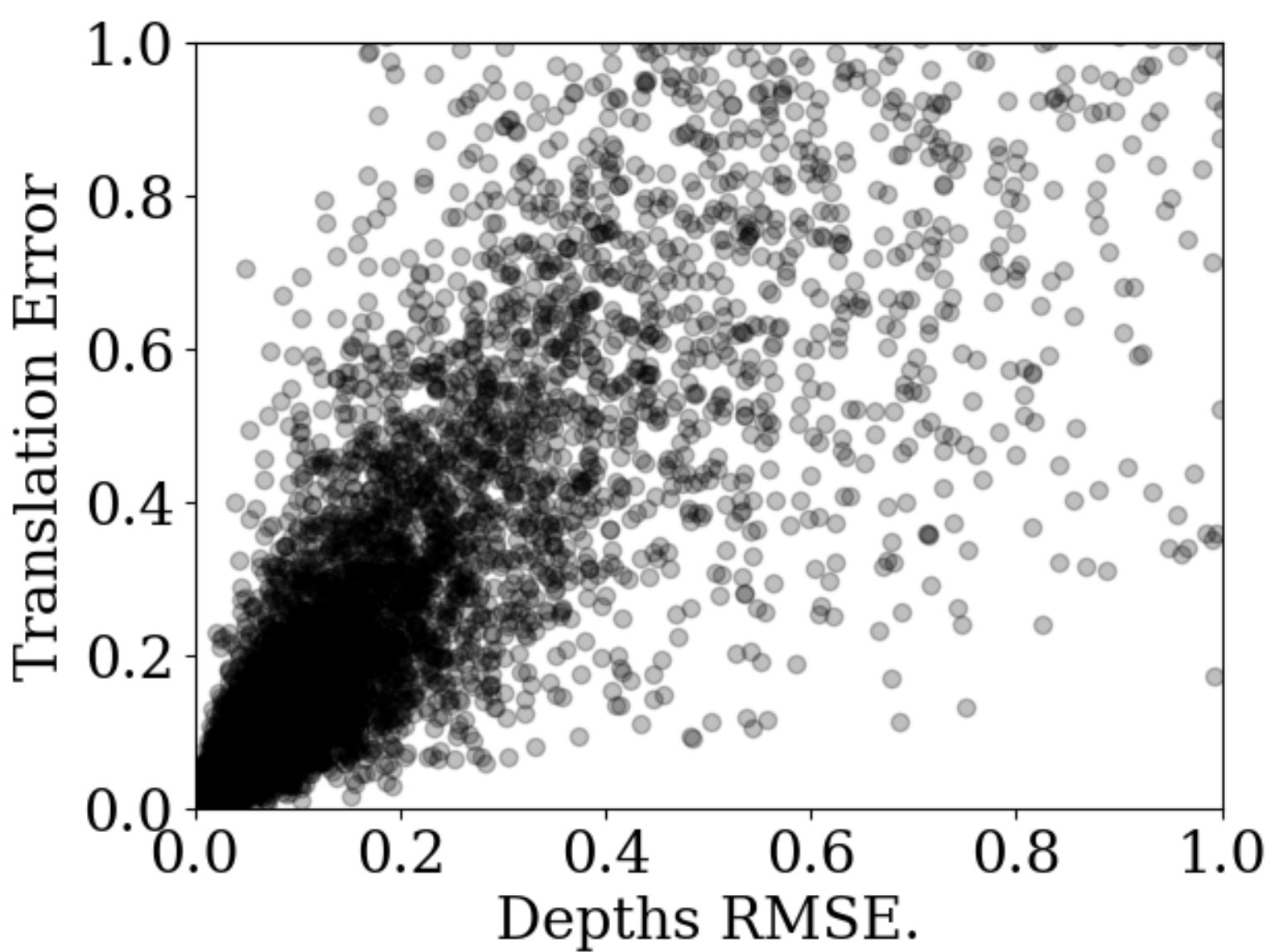}\\
\caption{{\bf Numerical Stability.} [Left to right] CDF of RMSE error distribution for estimated distances from $10^5$ trials and scatter plot of rotation and translation errors for gP4Pc from those trials.}
\label{fig:numerical_stability}
\end{figure}

\begin{figure}
\centering
\includegraphics[width=0.32\columnwidth]{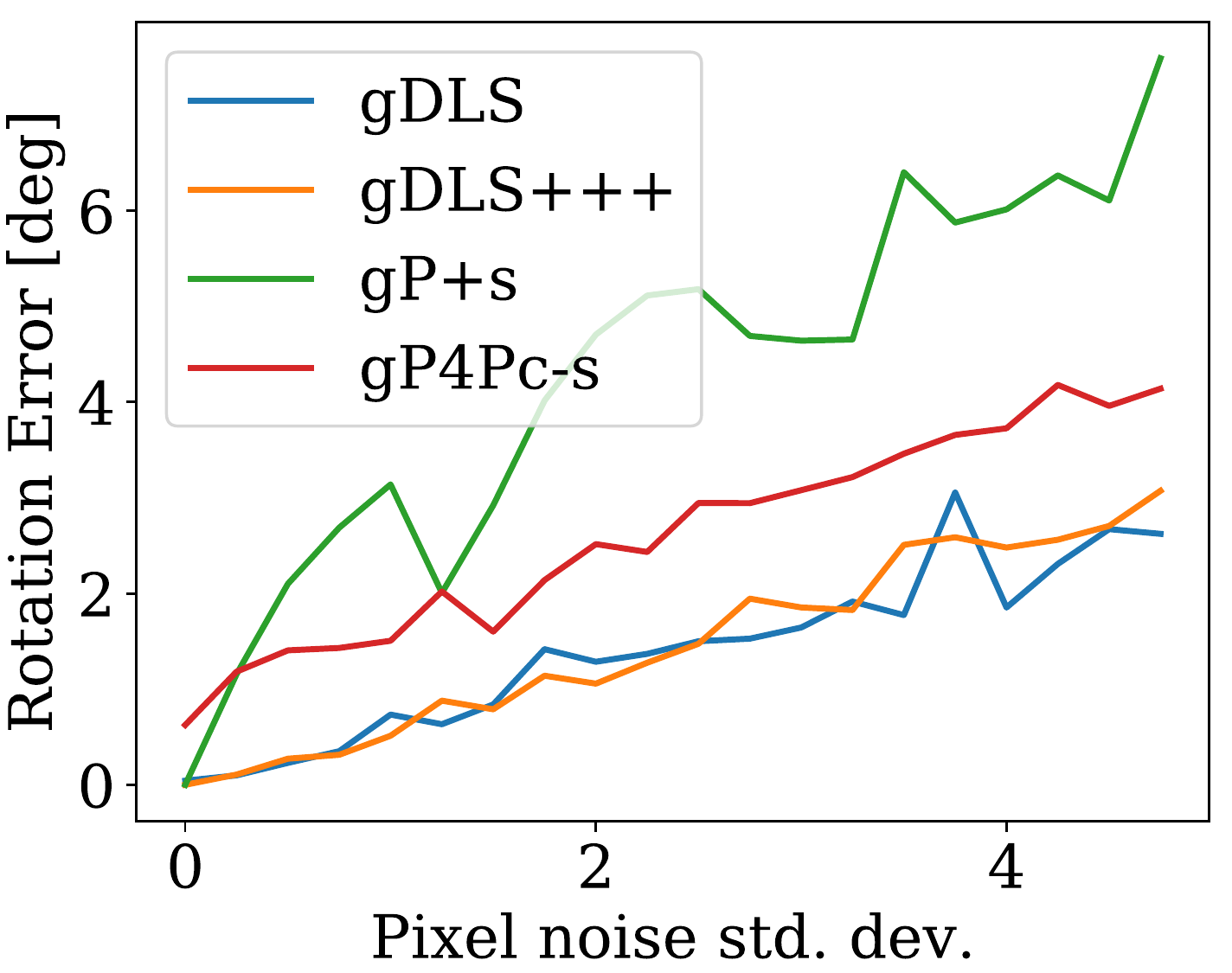}
\includegraphics[width=0.32\columnwidth]{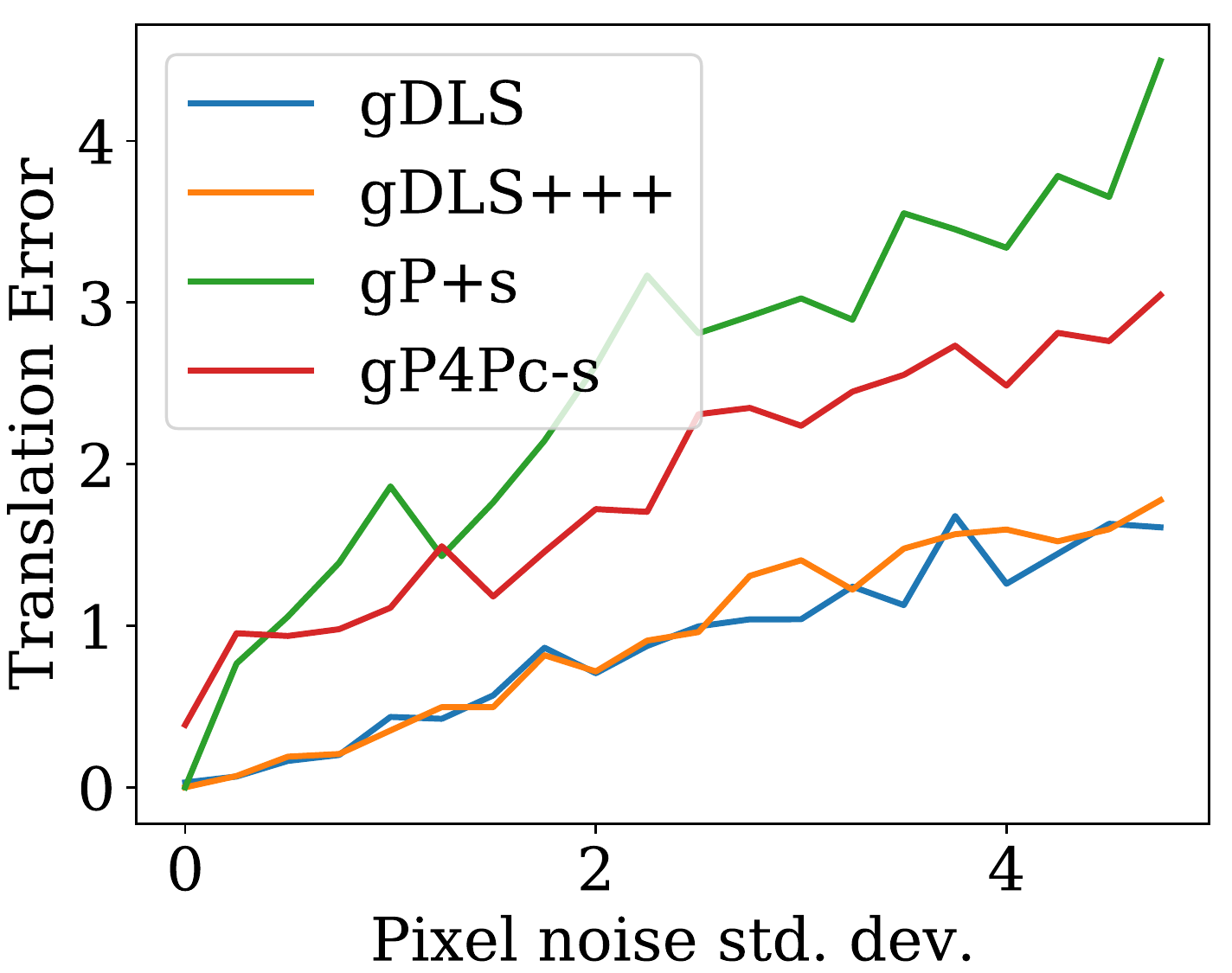}
\includegraphics[width=0.34\columnwidth]{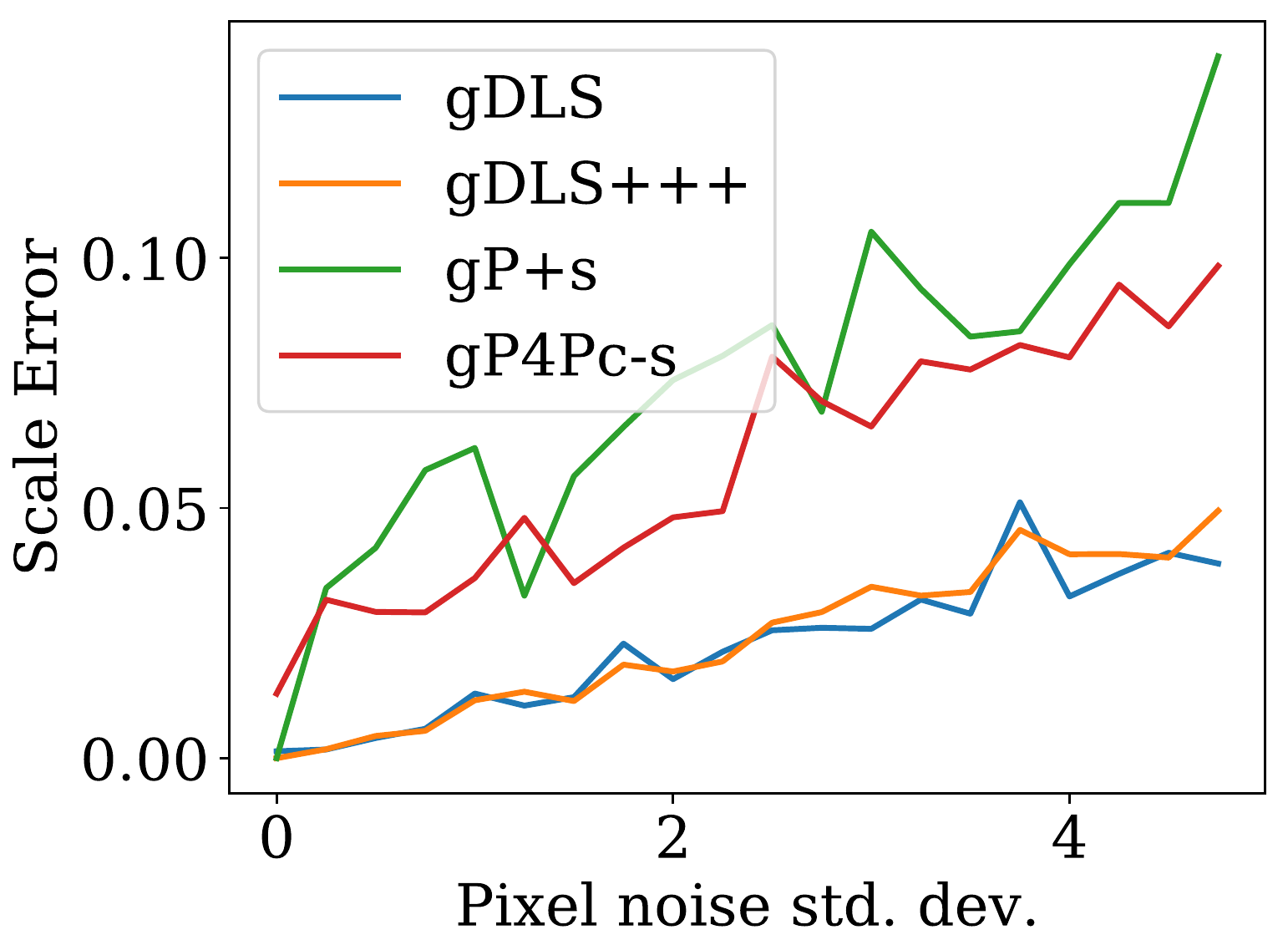}\\
\caption{{\bf Solver evaluation on synthetic data.} [Left to right] Rotation, translation and scale estimation errors for two gP4Pc variants and three baselines on minimal problems with noisy input.}
\label{fig:5def}
\vspace{-1mm}
\end{figure}

\noindent \textbf{Robust Estimation.}
For RANSAC-based robust pose estimation, we solve our minimal problem by first running  gP4Pc and then using Umeyama's method~\cite{umeyama1991least} for 3D point-to-point alignment in order to estimate the similarity transformation. We refer to that method as gP4Pc+s to differentiate it from another variant gP4Pc+a which we have experimented with. gP4Pc+a uses a linear method to compute an affine transformation from the four 3D point pairs instead of a similarity. During RANSAC, gP4Pc+a retains the best affine hypothesis \ie, the one with the most inliers and uses Umeyama's method at the end to compute the similarity transformation from all the inliers. However, we found that gP4Pc+s consistently outperforms gP4Pc+a.

\vspace{2mm}

\noindent \textbf{Implementation Details.} We implemented gP4Pc in C++ using the Theia library~\cite{sweeney2015theia} and the polynomial solver of Larsson~\etal\cite{larsson2018beyond}. We report the numerical stability, accuracy, and robustness of gP4Pc on synthetic and real data. The experiments include the following baselines
gP+s~\cite{ventura2014}, gDLS~\cite{sweeney2014}, and gDLS+++~\cite{sweeney2016}. We modified the gDLS implementation in Theia, specifically the routine that evaluates RANSAC hypotheses. Our version is faster than the original Theia implementation, and we used this modified routine for all the experiments. Finally, we also avoid the adaptive strategy to pick the number of RANSAC iterations used in Theia and instead use a fixed number of iterations.

\subsection{Evaluation on Synthetic Data}
\label{sec:synthetic_data}
In this section, we present results on four different experiments conducted on synthetic data.

\vspace{2mm}

\noindent \textbf{Numerical Stability}. We generated $10^5$ test problems, each with 100 points randomly selected from the $\left[-10, 10\right] \times \left[-10, 10\right] \times \left[-10, 10\right]$ cube and 10 random cameras in the $\left[-5, 5\right] \times \left[-5, 5\right] \times \left[10, 20\right]$ cube respectively, from which we sampled four point-ray pairs. No measurement noise was added. The true transformation in all cases was the identity transformation. Fig.~\ref{fig:numerical_stability} shows the cumulative distribution function (CDF) of the RMSE distribution for the distances from our solver and the scatter plots of the estimated rotation and translation errors. The estimated values are mostly accurate (RMSE $\leq 10^{-2}$) and produce accurate similarity estimates. However, sometimes erroneous distances (RMSE $\geq 1.0$) result in higher estimation errors.

\vspace{2mm}

\noindent \textbf{Evaluation on noisy data.} We compared gP4Pc+s (the similarity variant) and three baselines on synthetic problems where noise was added to the image measurements. Using a similar method as before, we generated 100 test problems with 10 cameras and 100 points each, with focal length set to 1000, image resolution set to 1000$\times$1000 and 3D points transformed by applying a randomly generated similarity transformation. The average estimation errors across the 100 runs are shown in Figure~\ref{fig:5def}. The experiment shows that gP4Pc is more accurate than gP+s but typically less accurate than gDLS and gDLS+++ on synthetic test data.

\vspace{2mm}

\noindent \textbf{Performance of solvers with RANSAC.} Next, we evaluated both gP4Pc solvers -- the '+s' and '+a' variants as well as the three baselines -- gP+s~\cite{ventura2014}, gDLS~\cite{sweeney2014} and gDLS+++~\cite{sweeney2016} on synthetic test problems containing outliers. We conducted three sets of runs with 25\%, 50\% and 75\% outliers respectively at different noise levels and in each case we used 1000 iterations of RANSAC and a
$2.5$ pixel inlier threshold. We present results for the 75\% outlier rate experiment in Figure~\ref{fig:synthetic_ransac}. The top half of the figure shows the average reprojection error, inlier count, rotation, translation and scale estimation errors for the five solvers. We observe that gP4Pc+s generally performs better than gP4Pc+a. Also, for most noise levels, the performance of gP4Pc+s is comparable to that of gDLS, gDLS+++ and gP+s although gDLS is a bit more accurate than all the other methods.

\vspace{2mm}

\noindent \textbf{Specialized solver for coplanar points.} We also evaluated our specialized solver for coplanar 3D points on synthetic data using a similar experiment setup, except that, in this case the 3D points lie on the $XY$ plane embedded in $\mathbb{R}^3$ and then a random rotation and translation is applied such that the points fall within the cube $\left[-10, 10\right] \times \left[-10, 10\right] \times \left[-10, 10\right]$.
The bottom half of Figure~\ref{fig:synthetic_ransac} shows the performance of the gP4Pc specialized solvers
and compares them to gDLS, gDLS+++ and gP+s. The most notable difference is in the running times.
See Table~\ref{tab:times} for the average running time for the minimal solvers.
For the general purpose case, gP+s is the fastest. However, for the case of coplanar points, our specialized solver is extremely fast. It takes 12.61 $\mu$-sec per problem, which is 3$\times$ faster than gP+s which take
42.61 $\mu$-sec per problem.

\begin{table}
\setlength\tabcolsep{3.0pt}
\centering
\label{tab:times}
\footnotesize{
\begin{tabular}{lccccc}
\toprule
Configuration & gP+s & gDLS & gDLS+++ & {\bf gP4Pc+s} & {\bf gP4Pc+a} \\
\midrule
Co-planar points &      42.9   & 597.6  & 223.9  & {\bf 12.61} &  19.19\\
General          & {\bf 41.25} & 578.29 & 224.18 &     723.66  & 706.48\\
\bottomrule\\
\end{tabular}
}
\caption{{\bf Timings.} in [$\mu$-sec] per minimal problem for the solvers. The two right entries in the top row are for our specialized solvers.}
\end{table}

\begin{table}
\setlength{\tabcolsep}{2.25pt} 
\begin{center}
\scriptsize{
\begin{tabular}{ccccccccc}
\toprule
~Seq.~ & \#Imgs.~ & ~{\bf gP4Pc+s (6p)}~ & ~{\bf gP4Pc+a (6p)}~ & ~{\bf gP4Pc+s (1p)}~ & ~{\bf gP4Pc+a (1p)}\\
\midrule
\multicolumn{6}{c}{Average Camera Position Estimate Error (in cm.)}\\
\midrule
1 & 8  & 6.30 $\pm$ 0.23   & 6.64 $\pm$ 0.87   & 6.24 $\pm$ 0.27   & 6.64 $\pm$ 0.73 \\
2 & 8  & 8.40 $\pm$ 0.18   & 8.44 $\pm$ 0.21   & 8.40 $\pm$ 0.18   & 8.41 $\pm$ 0.19 \\
3 & 32 & 7.34 $\pm$ 0.82   & 7.34 $\pm$ 0.76   & 7.42 $\pm$ 0.73   & 7.47 $\pm$ 0.78 \\
4 & 8  & 8.36 $\pm$ 0.42   & 8.42 $\pm$ 0.70   & 8.35 $\pm$ 0.44   & 8.48 $\pm$ 0.68 \\
5 & 14 & 6.39 $\pm$ 0.33   & 6.55 $\pm$ 0.45   & 6.39 $\pm$ 0.42   & 6.55 $\pm$ 0.44 \\
6 & 23 & 7.18 $\pm$ 0.28   & 7.23 $\pm$ 0.32   & 7.16 $\pm$ 0.26   & 7.21 $\pm$ 0.40 \\
7 & 8  & 7.43 $\pm$ 1.31   & 8.17 $\pm$ 1.73   & 7.47 $\pm$ 1.52   & 8.28 $\pm$ 1.95 \\
8 & 10 & 8.44 $\pm$ 0.91   & 8.38 $\pm$ 1.16   & 8.44 $\pm$ 0.89   & 8.45 $\pm$ 1.26 \\
9 & 6  & 6.73 $\pm$ 1.19   & 6.63 $\pm$ 1.44   & 6.54 $\pm$ 1.16   & 6.51 $\pm$ 1.40 \\
11& 57 & 7.02 $\pm$ 0.35   & 7.12 $\pm$ 0.39   & 7.05 $\pm$ 0.34   & 7.14 $\pm$ 0.38 \\
12& 66 & 5.85 $\pm$ 0.92   & 6.51 $\pm$ 1.08   & 5.88 $\pm$ 0.92   & 6.08 $\pm$ 1.22 \\
\midrule
\multicolumn{2}{c}{(mean $\pm$ s.d.)} & 7.22 $\pm$ 1.13 & 7.40 $\pm$ 1.22 & \textbf{7.21 $\pm$ 1.15} & 7.38 $\pm$ 1.31 \\
\midrule
\multicolumn{2}{c}{time/m.p. (ms.)} & 0.249 &  0.250 &  0.237 &  0.238 \\
\midrule
\multicolumn{2}{c}{time (sec.)} & 10.35 &  13.40 &  1.729 &  2.243 \\
\bottomrule
\end{tabular}
}
\end{center}
\label{tab:ablation_study}
\caption{\textbf{Ablation Study - Office sequences:} The camera position errors (mean, standard dev. in cm.) per sequence for the four proposed variants of our method. The variants using one and six permutations are indicated as (1p) and (6p) respectively.
The variants using similarity and affine transforms are indicated as '+a' and '+s' respectively. Timings are reported
for just the minimal solver as well as for the total time with 1000 RANSAC iterations.}
\end{table}

\begin{table}
\setlength{\tabcolsep}{2.6pt} 
\begin{center}
\scriptsize{
\begin{tabular}{ccccccccc}
\toprule
~~Seq.~~ & ~~\#Imgs.~~ & ~~gP+s~~~~~~~ & ~~~~~~~gDLS~~~~~~ & ~~~~~~gDLS+++~~~ & ~gP4Pc~\\
\midrule
\multicolumn{6}{c}{Average Camera Position Estimate Error (in cm.)}\\
\midrule
1 & 8   & 6.24 $\pm$ 0.25          & ~6.12 $\pm$ 0.27          & 6.24 $\pm$ 0.23          & 6.24 $\pm$ 0.27 \\
2 & 8   & 8.39 $\pm$ 0.14          & ~8.42 $\pm$ 0.14          & 8.39 $\pm$ 0.14          & 8.40 $\pm$ 0.18 \\
3 & 32  & 7.26 $\pm$ 0.74          & ~7.31 $\pm$ 0.70          & 7.46 $\pm$ 0.80          & 7.42 $\pm$ 0.73 \\
4 & 8   & 8.39 $\pm$ 0.47          & ~8.41 $\pm$ 0.41          & 8.33 $\pm$ 0.43          & 8.35 $\pm$ 0.44 \\
5 & 14  & 6.30 $\pm$ 0.22          & ~6.29 $\pm$ 0.21          & 6.31 $\pm$ 0.22          & 6.39 $\pm$ 0.42 \\
6 & 23  & 7.20 $\pm$ 0.23          & ~7.15 $\pm$ 0.24          & 7.17 $\pm$ 0.29          & 7.16 $\pm$ 0.26 \\
7 & 8   & 7.54 $\pm$ 1.34          & ~7.42 $\pm$ 1.24          & 7.15 $\pm$ 1.07          & 7.47 $\pm$ 1.52 \\
8 & 10  & 8.51 $\pm$ 0.80          & ~8.42 $\pm$ 0.75          & 8.40 $\pm$ 0.86          & 8.44 $\pm$ 0.89 \\
9 & 6   & 6.51 $\pm$ 1.08          & ~6.64 $\pm$ 1.05          & 6.77 $\pm$ 1.16          & 6.54 $\pm$ 1.16 \\
11 & 57 & 7.05 $\pm$ 0.31          & ~6.99 $\pm$ 0.30          & 6.99 $\pm$ 0.28          & 7.05 $\pm$ 0.34 \\
12 & 66 & 5.59 $\pm$ 0.95          & ~5.57 $\pm$ 1.00          & 5.63 $\pm$ 0.99          & 5.88 $\pm$ 0.92 \\
\midrule
\multicolumn{2}{c}{(mean $\pm$ s.d.)} & 7.18 $\pm$ 1.17 & 7.16 $\pm$ 1.15 & 7.17 $\pm$ 1.13 & 7.21 $\pm$ 1.15 \\
\midrule
\multicolumn{2}{c}{time/m.p. (ms.)} & 0.028 & 0.199 &  0.102 &  0.237 \\
\midrule
\multicolumn{2}{c}{num solns./m.p.} & 4.8 & 2.5 &  2.7 &  2.6 \\
\midrule
\multicolumn{2}{c}{time (sec.)} & 2.279 & 1.719 &  1.431 &  1.729 \\
\bottomrule
\end{tabular}
}
\end{center}
\label{tab:ventura_results}
\caption{\textbf{Quantitative Evaluation - Office sequences:} The camera position errors (mean, standard dev. in cm.) per sequence for our method gP4Pc+s and baselines (gP+s, gDLS, gDLS+++). All the methods have similar performance on this dataset.
The average running time for the minimal solvers and the total time for pose estimatin using 1000 RANSAC iterations are reported as well as the average number of valid solutions from the minimal solvers.}
\end{table}

\begin{table*}
\setlength{\tabcolsep}{2.4pt}
    \centering
    \scriptsize {
    \begin{tabular}{l ccccc c ccccc c ccccc c ccccc}
        \toprule
        \cmidrule{1-24}
        & \multicolumn{5}{c}{Rotation Error $\left[\text{deg}\right] ~ (10^{-2})$} & & \multicolumn{5}{c}{Translation Error ~ $(10^{-3})$} & & \multicolumn{5}{c}{Scale Error~ $\left(10^{-4}\right)$} & & \multicolumn{5}{c}{$\text{time} ~ \left[ \text{sec} \right]$}\\
        \cmidrule{2-6} \cmidrule{8-12} \cmidrule{14-18} \cmidrule{20-24}
        &
        gP+s & ~gDLS & gDLS\tiny{+++} & 3Q3~ & gP4Pc & &
        gP+s & ~gDLS & gDLS\tiny{+++} & 3Q3~ & gP4Pc & &
        gP+s & ~gDLS & gDLS\tiny{+++} & 3Q3~ & gP4Pc & &
		gP+s & ~gDLS & gDLS\tiny{+++} & 3Q3~ & gP4Pc  \\
        \cmidrule{1-24}
        $T_1$ &
        10.8 & 14.1 & 14.0 & 13.5 & 14.0 & &
        7.91 & 1.11 & 1.10 & 28.4 & 1.10 & &
        5.01 & 8.54 & 8.61 & 19.0 & 8.61 & &
        336 & 195 & 189 & 162 & 189 \\
        $T_2$ &
        10.6 & 10.5 & 10.6 & 11.1 & 11.4 & &
        20.5 & 19.7 & 19.9 & 21.7 & 21.7 & &
        14.1 & 13.4 & 12.6 & 17.5 & 16.2 & &
        155 & 84 & 82 & 72 & 96 \\
        $T_3$ &
        9.69 & 8.06 & 7.68 & 10.2 & 9.27 & &
        68.7 & 57.9 & 54.6 & 71.8 & 68.4 & &
        58.3 & 51.0 & 51.3 & 59.3 & 60.7 & &
        153 & 74 & 74 & 68 & 68 \\
        $T_4$ &
        9.04 & 8.49 & 8.51 & 10.2 & 9.10 & &
        12.6 & 12.2 & 12.7 & 15.0 & 12.8 & &
        16.2 & 15.5 & 16.0 & 17.6 & 15.2 & &
        302 & 150 & 152 & 138 & 167\\
        $T_5$ &
        5.20 & 5.02 & 5.20 & 5.97 & 5.69 & &
        9.59 & 9.34 & 9.59 & 11.4 & 11.5 & &
        21.6 & 21.8 & 21.6 & 28.5 & 32.1 & &
        362 & 333 & 362 & 295 & 282 \\
        $T_6$ &
        7.72 & 6.09 & 6.29 & 7.00 & 6.77 & &
        18.0 & 16.7 & 16.5 & 16.9 & 16.0 & &
        91.3 & 76.2 & 80.0 & 104.0 & 89.7 & &
        115 & 47 & 54 & 53 & 43\\
        \cmidrule{1-24}
        $T_{avg}$ & 8.84	& 8.71	& 8.71 & 9.7 & 9.37 & &
        22.9 & 19.5 & 19.1 & 27.5 & 21.9 & &
        34.4 & 31.1 & 31.7 & 41 & 37.1 & &
        237 & 147 & 152 & 131.1 & 141	\\
        \cmidrule{1-24}
        $K_1$ &
         1.66 & 1.56 & 1.64 & 1.80 & 1.58 & &
         1.34 & 1.25 & 1.29 & 1.43 & 1.29 & &
         2.45 & 2.56 & 2.70 & 2.40 & 2.69 & &
         2.03 & 1.34 & 1.44 & 0.94 & 0.89 \\
        $K_2$ &
         7.42 & 7.62 & 7.69 & 8.32 & 7.57 & &
         7.61 & 6.91 & 7.07 & 8.50 & 7.30 & &
         0.97 & 0.87 & 0.98 & 107.0& 0.90 & &
         2.95 & 1.79 & 1.95 & 1.34 & 1.90 \\
        $K_3$ &
         10.2 & 10.1 & 10.1 & 13.1 & 11.6 & &
         6.01 & 6.01 & 5.93 & 7.73 & 7.00 & &
         46.2 & 40.6 & 43.1 & 50.2 & 47.0 & &
         20.2 & 11.8 & 12.2 & 8.50 & 12.8 \\
        $K_4$ &
         2.67 & 2.58 & 2.58 & 3.24 & 3.05 & &
         3.45 & 3.34 & 3.54 & 3.80 & 3.39 & &
         11.1 & 11.2 & 11.8 & 13.6 & 12.2 & &
         2.46 & 1.61 & 1.41 & 1.08 & 1.67 \\
        $K_5$ &
         2.99 & 2.77 & 2.79 & 2.72 & 2.47 & &
         1.54 & 1.44 & 1.45 & 1.39 & 1.43 & &
         7.05 & 7.61 & 7.45 & 6.93 & 7.50 & &
         1.80 & 1.31 & 1.13 & 0.77 & 1.08 \\
        $K_6$ &
         6.11 & 5.76 & 6.50 & 6.51 & 6.84 & &
         4.92 & 4.70 & 4.96 & 4.84 & 5.26 & &
         23.8 & 21.8 & 24.9 & 22.7 & 23.6 & &
         7.30 & 4.16 & 4.71 & 3.36 & 3.72  \\
         \cmidrule{1-24}
        $K_{avg}$ & 5.18 & 5.07 & 5.22 & 5.95 & 5.52 & &
         4.15 & 3.94 & 4.04 & 4.62 & 4.27 & &
         15.3 & 14.1 & 15.2 & 34.0 & 15.7 & &
         6.12 & 3.67 & 3.81 & 2.62 & 3.68 \\
        \bottomrule\\
    \end{tabular}
}
\caption{\textbf{Results on TUM \& KITTI:}
See upper and lower sections for results on the TUM sequences $T_{1:6}$ are \textit{(Fr1 Desk, Fr1 Desk2, Fr2 LargeNoLoop, Fr1 Room, Fr2 Pion.SLAM and Fr2 Pion.SLAM 2)} and the KITTI sequences $K_{1:6}$ \textit{(Drive 1, 9, 19, 22, 23 and 29)} respectively. gDLS and gDLS+++ lead in terms of accuracy followed by gP4Pc. 3Q3 is the fastest method followed by gP4Pc in terms of total running time. But, gP4Pc is more accurate than 3Q3 and only slightly slower on average (it is actually faster on 3 of the 12 sequences).}
\label{tab:tum_kitti_results}
\vspace{-2mm}
\end{table*}

\subsection{Evaluation on Office Dataset}
\label{sec:ventura}

In this section, we report results on real datasets using all the solvers for a SLAM-trajectory registration task. This task requires localizing a moving camera \wrt an existing 3D reconstruction by solving the generalized gP+s problem. We will first report results on the Office dataset~\cite{ventura2014}.
This dataset contains 12 sequences, each of which contains ground truth camera poses for keyframes that were collected using an ART-2 optical tracker, the SfM scene reconstruction (3D point cloud and 2D image measurements and associated SIFT features~\cite{lowe1999object}). We discarded sequence 10 as it had invalid data. To obtain 2D--3D correspondences, we used exhaustive nearest neighbor SIFT descriptor matching and the standard Lowe's ratio test with a threshold of $0.7$. In our RANSAC implementation, we used a reprojection error threshold of $2$ pixels and 1000 iterations and we did not run any nonlinear pose optimization. We repeated each experiment 100 times and report average errors.

\begin{figure}
\centering
\includegraphics[width=\columnwidth]{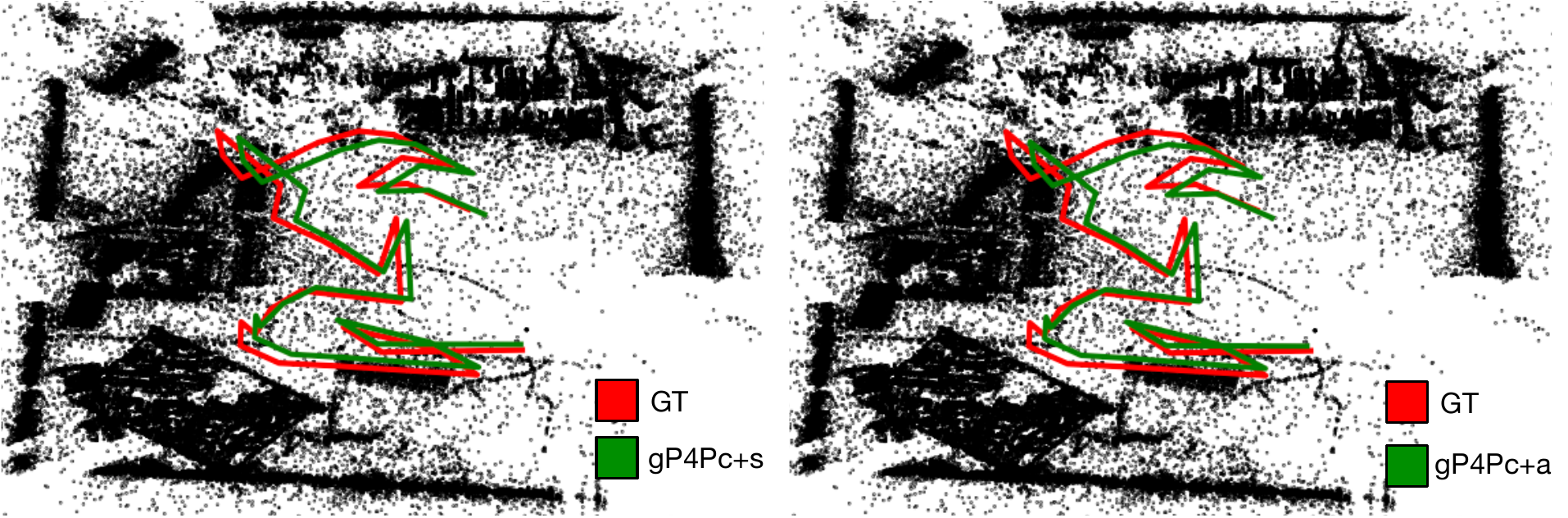}
\caption{Top view of the estimated and ground truth trajectories for gP4Pc+s (on the left) and gP4Pc+a (on the right).}
\label{results_real}
\vspace{-3mm}
\end{figure}

We first report results from an ablation study for our solver in Table~\ref{tab:ablation_study}. The mean camera position error and the standard deviation in cm. from 100 runs is shown for each sequence. We compare two variants each of gP4Pc+s and gP4Pc+a, each with either one or six random permutations respectively (indicated by suffices (1p) and (6p)). We observe that gP4Pc+s (1p) has the smallest error across all sequences and has a mean error of 7.21cm pixels and a standard deviation of 1.15cm. Fig.~\ref{results_real} shows trajectories computed by gP4Pc+a (1p) and gP4Pc+s (1p) for Sequence 3.

Next we compared gP4Pc+s(1p) with gP+s, gDLS and gDLS+++ on the Office sequences. The results are shown in Table~\ref{tab:ventura_results}. In the rest of the paper, we drop the (1p) suffix for brevity. While gP4Pc had a mean position error of 7.21 $\pm$ 1.15 cm, gDLS with a mean error of 7.16 $\pm$ 1.15cm was the best method. However, notice that the error margin between the four methods was extremely small considering the standard deviation of their errors. Therefore, we conclude that gP4Pc is competitive with the state of the art on this dataset. The mean running time of the solver is also shown in the table as well as the average number of hypotheses and the total running times. Even though gP+s had the fastest solver, it produces many more hypotheses to verify whereas both gDLS and gP4Pc produce fewer hypotheses. The total running time for gDLS+++ was 1.43 making it the fastest but gP4Pc and gDLS take 1.73 and 1.72 seconds respectively and are not very far behind in terms of running times.

\subsection{Evaluation on TUM and KITTI datasets}
\label{sec:tumkitti}
We also evaluated gP4Pc on the TUM~\cite{sturm2012benchmark} and KITTI~\cite{geiger2013vision} datasets using the protocol proposed by Fragoso~\etal~\cite{fragoso2020}. They reconstruct the scene from the SLAM trajectories and remove a series of frames from the reconstructions to form a generalized camera query and apply a random similarity transformation to the query camera poses. This makes it possible to calculate rotation, position and scale errors, unlike the Office dataset~\cite{ventura2014} where only ground truth camera position data is available. We compared gP4Pc with gP+s~\cite{ventura2014}, gDLS~\cite{sweeney2014}, 3Q3~\cite{kukelova2016} and gDLS+++~\cite{sweeney2016}. We excluded gDLS*~\cite{fragoso2020} which becomes identical to gDLS+++ in the unconstrained setting \ie when pose priors are absent.

We report average errors across 100 runs of the same query. Table~\ref{tab:tum_kitti_results} shows the rotation, translation, scale errors and timings for the five methods. The upper and bottom parts of the Table presents results on the TUM and KITTI datasets respectively. The rotation, translation, and scale errors for gP4Pc are slightly larger than that of gDLS and gDLS+++. While 3Q3 is the fastest method on average, it is also the least accurate. The experiment confirms that gP4Pc produces competitive pose-and-scale estimates. In terms of the total running time with RANSAC, gP4Pc is only slightly slower than 3Q3 on average (in fact, it was faster than 3Q3 in three cases). gDLS (with our modified implementation) and gDLS+++ are usually ranked next in terms of speed followed by gP+s. The reported timings were obtained on a PC with a 2.1 GHz Xeon CPU with 48 GB RAM using single threaded implementations. We analyzed the number of hypotheses generated by different minimal solvers on the ``Drive 1" sequence, which for gP+s, gDLS, gDLS+++, 3Q3 and gP4Pc is about 3.9, 1.6, 2.4, 1.9 and 1.1 respectively. Thus, on these datasets, gP4Pc tends to produce the fewest solutions to evaluate during RANSAC resulting in a noticeable speedup compared to other methods. 
\vspace{-1mm}
\section{Conclusions}
We presented gP4Pc, a new method for generalized pose-and-scale estimation
from four point-ray pairs. Unlike existing methods (\eg, gDLS~\cite{sweeney2014}, gDLS+++~\cite{sweeney2016}, and gP+s~\cite{ventura2014}) that use a least-squares-based formulation, gP4Pc instead uses 4-point congruence constraints to solve the problem. By solving a new polynomial system using an efficient Gr{\"o}bner-based polynomial solver, we estimate the point distances along the rays from the pinholes and then estimate the final transformation using 3D point-point alignment methods. We also presented a fast version of gP4Pc for when the 3D points are coplanar.  Our experiments show that gP4Pc is comparable to the state-of-the-art solvers in terms of accuracy. On the TUM and KITTI datasets, it often generated fewer solution candidates that needed to be checked compared to existing methods and provided a good trade-off between accuracy and speed.

Currently, the order in which the point-ray pairs are presented to the solver matters in certain geometric configurations. Currently, we do not have an effective strategy to select the optimal order or permutation of the input and rely on randomization. However if one could efficiently find a good order, it could potentially boost the accuracy of the method further without adding any computational overhead. This could be an interesting direction for future work. 

\appendix

\section{Specialized Method for Coplanar Points}
\label{sec:coplanar}

Recall, that we are given four rays, parameterized using four 3D points, $\pp_1$, $\pp_2$, $\pp_3$, $\pp_4$, denoting the projection centers and four 3D unit vectors, $\uu_1$, $\uu_2$, $\uu_3$, $\uu_4$, denoting the ray directions. We are also given four coplanar 3D points, $\xx_1$, $\xx_2$, $\xx_3$ and $\xx_4$
from which we get ratios $r_1$, $r_2$, $K_{1213}$ and $K_{1234}$ (see paper for the details). We need to compute four unknown scalar values, $s_1$, $s_2$, $s_3$, $s_4$, such that
\begin{align}
\yy_1 = \pp_1 + s_1 \uu_1\\
\yy_2 = \pp_2 + s_2 \uu_2\\
\yy_3 = \pp_3 + s_3 \uu_3\\
\yy_4 = \pp_4 + s_4 \uu_4
\end{align}
and each $\yy_i = T \xx_i$ for $i=1\ldots4$ where $T$ is an unknown 3D similarity transformation.

Now, recall that in the main paper, for the coplanar points case, we presented three linear constraints and one quadratic constraint in $s_1$, $s_2$, $s_3$, $s_4$. These equations are repeated here for convenience (see Equations~\ref{eq1} and~\ref{eq1b} respectively):
\begin{align}
\nonumber
(1 - r_1) (\pp_1 + s_1 \uu_1) + r_1 (\pp_2 + s_2 \uu_2) = \\
(1 - r_2) (\pp_3 + s_3 \uu_3) + r_2 (\pp_4 + s_4 \uu_4).
\label{eq1}
\end{align}

\noindent The quadratic equation is
\begin{align}
(\boldsymbol{\beta}_{12} - K_{1213} \boldsymbol{\beta}_{13})^\intercal \ss = 0,
\label{eq1b}
\end{align}
where $\ss = \\ \left[s_1^2 \: s_2^2 \: s_3^2 \: s_4^2 \: s_1s_2 \: s_1s_3 \: s_1s_4 \: s_2s_3 \: s_2s_4 \: s_3s_4 \: s_1 \: s_2 \: s_3 \: s_4 \: 1\right]^\intercal $.

In order to derive our specialied method for coplanar points, we treated $s_1$, $s_2$, $s_3$ as variables and rewrote the three linear equations in Equation~\ref{eq1} in matrix form as follows:
\begin{align}
\begin{bmatrix}
a_1 & b_1 & c_1 \\ a_2 & b_2 & c_2 \\ a_3 & b_3 & c_3
\end{bmatrix}
\begin{bmatrix}
s_1 \\ s_2 \\ s_3
\end{bmatrix}
=
\begin{bmatrix}
d_1 \\ d_2 \\ d_3
\end{bmatrix}.
\label{eq22}
\end{align}
In Equation~\ref{eq22}, the terms $d_1$, $d_2$ and $d_3$ are linear functions of $s_4$.
We then consider closed form solutions of the linear system in Equation~\ref{eq22} based on Cramer's Rule.
Thus, we obtain closed form expressions for $s_1$, $s_2$ and $s_3$ in terms of $s_4$, which take the following form:
\begin{align}
s_1 &= G_1 s_4 + H_1 \label{eq2a}\\
s_2 &= G_2 s_4 + H_2 \label{eq2b}\\
s_3 &= G_3 s_4 + H_3 \label{eq2c}.
\end{align}
Although, we have omitted the expressions for the six terms
$G_1$, $H_1$, $G_2$, $H_2$, $G_3$ and $H_3$ here, they can be
derived using basic algebraic manipulation and then plugging in all the
values into the formulae for Cramer's rule.
The next step is to substitute the expressions of $s_1$, $s_2$ and
$s_3$ from Equation~\ref{eq2a},~\ref{eq2b} and~\ref{eq2c} into
the quadratic equation in Equation~\ref{eq1b}. This gives us a new quadratic equation in $s_4$:

\begin{align}
A s_4^2 + B s_4 + C = 0
\label{quadratic}
\end{align}
where, the coefficients $A$, $B$ and $C$ are defined as follows:

\begin{align}
A\,=&\, C_1 G_1^2 + C_2 G_2^2 + C_3 G_3^2 + C_4 G_1 G_2 + C_5 G_1 G_3\\
\nonumber B\,=&\, 2\,(C_1 G_1 H_1 + C_2 G_2 H_2 + C_3 G_3 H_3)\\
\nonumber   & + C_4 (G_1 H_2 + G_2 H_1) + C_5 (G_1 H_3 + G_3 H_1)\\
            & + C_6 G_1 + C_7 G_2 + C_8 G_3\\
\nonumber \\
\nonumber C\,=&\, C_1 H_1^2 + C_2 H_2^2 + C_3 H_3^2 + C_4 H_1 H_2 + C_5 H_1 H_3\\
   &  + C_6 H_1 + C_7 H_2 + C_8 H_3 + C_9.
\label{eq5}
\end{align}


In the above expressions for $A$, $B$ and $C$, the terms $C_k$ for $k=1\ldots9$ are defined as follows:
\begin{align}
C_1 \,=&\, 1 - K_{1213}\\
C_2 \,=&\, 1\\
C_3 \,=&\, - K_{1213}\\
C_4 \,=&\, -2 \cdot \uu_1^\intercal \uu_2\\
C_5 \,=&\, 2 \cdot K_{1213} \cdot \uu_1^\intercal \uu_3\\
C_6 \,=&\, 2 \cdot \uu_1^\intercal \pp_6 - 2 \cdot K_{1213} \cdot \uu_1^\intercal \pp_8\\
C_7 \,=&\, -2 \cdot \uu_2^\intercal \pp_6\\
C_8 \,=&\, 2 \cdot K_{1213} \cdot \uu_3^\intercal \pp_8\\
C_9 \,=&\, (\pp_6^\intercal \pp_6) - K_{1213} \cdot (\pp_8^\intercal \pp_8)
\label{eq6}
\end{align}
where $\pp_6 = \pp_1 - \pp_2$ and $\pp_8 = \pp_1 - \pp_3$. The next step in our method is to find the roots of the quadratic equation shown in Equation~\ref{quadratic}. We then check for positive values for $s_4$ and then substitute them into Equations~\ref{eq2a},~\ref{eq2b} and~\ref{eq2c} to get values of $s_1$, $s_2$ and $s_3$. We return solutions where these three values are also positive.

\section{Polynomial Solver}
\label{sec:general_poly_solver}

As explained in Section 3 in the main paper, we need to solve a polynomial system consisting of four quadratic polynomials in $\mathbf{s}$. To obtain a solver for this polynomial system, we used autogen, the automatic Gr{\"o}bner-based polynomial solver generator from Larsson~\etal~\cite{larsson2018beyond}.
While autogen can generate problem instances with random integer coefficients, we found that such instances were not representative of the geometry underlying our polynomial system and the generated solver produced inaccurate results. To address this issue, we generated our own synthetic problem instances using the following steps.

The polynomial system stated in Equations (16) and (17) in the main paper can be arranged in matrix form, $A\cdot\mathbf{s}=0$, where the matrix $A \in \mathbb{R}^{4 \times 15}$ holds the coefficients of the polynomial system. We generated synthetic problem instances, following the protocol described in the main paper. Given $K$ problem instances and their respective coefficient matrices $A_k, \forall k=1,\hdots,K$, we computed the average coefficient matrix $A_{\text{avg}} = \frac{1}{l}\sum_{i=1}^K A_k$ and then multiplied every element of the
matrix $A_{\text{avg}}$ by a scale factor $S$ before rounding the nonzero entries to the nearest integers.
We set $K$ and $S$ to the following values: $K=100$ and $S=50$.

We observed that the value of the random coefficients of the polynomial system generated in this way, appeared to be representative of the underlying problem. All nonzero entries in the resulting problem instance were integers and the autogen package produced a stable and accurate solver from it. The obtained polynomial solver uses an elimination template matrix with 97 rows and 113 columns and a $16 \times 16$ action matrix. The polynomial solver returns up to 16 solutions and we only keep those that are real and positive.

\footnotesize {
\bibliographystyle{ieee}
\bibliography{main}
}
\end{document}